\documentclass{article} 
\usepackage[]{iclr2022_conference,times}

\usepackage{amsmath,amsfonts,bm}

\def\eqref#1{equation~\ref{#1}}

\def\1{\bm{1}}

\DeclareMathAlphabet{\mathsfit}{\encodingdefault}{\sfdefault}{m}{sl}
\SetMathAlphabet{\mathsfit}{bold}{\encodingdefault}{\sfdefault}{bx}{n}

\usepackage{hyperref}
\usepackage{url}
\usepackage{authblk}
\makeatletter
\renewcommand\AB@affilsepx{\protect\Affilfont}
\makeatother
\usepackage[symbol]{footmisc}

\usepackage{amsmath}
\usepackage[pdftex]{graphicx}
\usepackage{algpseudocode}
\usepackage[title]{appendix}
\usepackage{tabularx}
\usepackage{soul}
\usepackage{diagbox}
\usepackage{multirow}
\usepackage{algorithm,algpseudocode,float}
\usepackage{lipsum}
\usepackage{amsthm}
\usepackage[utf8]{inputenc} 
\usepackage[T1]{fontenc}    
\definecolor{mycolor}{RGB}{0,20,115}
\hypersetup{
	colorlinks=true,
	linkcolor=mycolor,
	filecolor=magenta,      
	urlcolor=mycolor,
	citecolor=mycolor
}
\usepackage{booktabs}       
\usepackage{amsfonts}       
\usepackage{nicefrac}       
\usepackage{microtype}      
\usepackage{wrapfig}
\usepackage{subcaption}

\makeatletter
\newcommand{\multiline}[1]{
	\begin{tabularx}{\dimexpr\linewidth-\ALG@thistlm}[t]{@{}X@{}}
		#1
	\end{tabularx}
}
\makeatother

\makeatletter

\title{Batch-ensemble Stochastic Neural Networks for Out-of-Distribution Detection}

\author[1]{\textbf{Xiongjie Chen}}
\author[1]{\textbf{Yunpeng Li}}
\author[2]{\textbf{Yongxin Yang}}
\affil[1]{University of Surrey,\;}
\affil[2]{University of Edinburgh,\;}
{
	\makeatletter
	\renewcommand\AB@affilsepx{\protect\\\Affilfont}
	\makeatother
	
	\affil[ ]{}
	
	\makeatletter
	\renewcommand\AB@affilsepx{\protect\Affilfont}
	\makeatother
	
	\affil[1]{\texttt{\{xiongjie.chen, yunpeng.li\}@surrey.ac.uk},\;}
	\affil[2]{\texttt{yongxin.yang@ed.ac.uk}}
}

\graphicspath{{Figures/}}
\iclrfinalcopy 
\begin{document}

	\maketitle
	
	\def\thickhline{%
		\noalign{\ifnum0=`}\fi\hrule \@height \thickarrayrulewidth \futurelet
		\reserved@a\@xthickhline}
	\def\@xthickhline{\ifx\reserved@a\thickhline
		\vskip\doublerulesep
		\vskip-\thickarrayrulewidth
		\fi
		\ifnum0=`{\fi}}
	\makeatother
	\newlength{\thickarrayrulewidth}
	\setlength{\thickarrayrulewidth}{3\arrayrulewidth}
	\vspace{-2em}
	
	\begin{abstract}
		Out-of-distribution (OOD) detection has recently received much attention from the machine learning community due to its importance in deploying machine learning models in real-world applications. In this paper we propose an uncertainty quantification approach by modelling the distribution of features. We further incorporate an efficient ensemble mechanism, namely batch-ensemble, to construct the batch-ensemble stochastic neural networks (BE-SNNs) and overcome the feature collapse problem. We compare the performance of the proposed BE-SNNs with the other state-of-the-art approaches and show that BE-SNNs yield superior performance on several OOD benchmarks, such as the Two-Moons dataset, the FashionMNIST vs MNIST dataset, FashionMNIST vs NotMNIST dataset, and the CIFAR10 vs SVHN dataset.
	\end{abstract}
	
	\section{Introduction}
	
	\label{sec:introduction}
	Identifying out-of-distribution (OOD) samples that neural networks have never observed on is a critical and challenging task for applying deep learning models in real-world scenarios. Various OOD detection methods have been proposed to enhance the reliability of deep learning models~\citep{wen2019batchensemble,ren2019likelihood,zhou2021step,sun2021react,liu2020energy,van2020uncertainty,wan2018rethinking,kim2021locally}. One main branch of current OOD detection approaches is the so-called post hoc methods, where some selected statistics of the model outputs are post-processed after training, in order to differentiate in-distribution samples from OOD samples~\citep{ren2019likelihood,zhou2021step,sun2021react,liu2020energy}. In contrast, another line of work resorts to evaluating the uncertainties of trained models in their predictions to detect OOD samples~\citep{wen2019batchensemble,van2020uncertainty,wan2018rethinking}. In this paper, we focus on the latter one, i.e. detecting OOD samples using uncertainty estimation techniques.
	
	A majority of approaches for estimating uncertainty in deep learning models can be categorized into two classes. The first class relies on the ensemble of deep neural networks~\citep{lakshminarayanan2017simple,wen2019batchensemble}, where the outputs of multiple individually trained models are combined to estimate the uncertainty. The second family of uncertainty estimation approaches aims to measure the predictive uncertainty using deterministic single forward pass neural networks~\citep{van2020uncertainty,wan2018rethinking,mukhoti2021deep}, where the uncertainty is estimated by modelling the distribution of features.
	
	Despite their success on benchmark datasets, both deep ensemble methods and deterministic single forward pass neural networks methods have limitations respectively. An obvious disadvantage of deep ensemble methods is their computational costs~\citep{durasov2021masksembles,wen2019batchensemble,dusenberry2020efficient}. Particularly, ensemble methods are limited in practice since each ensemble member requires an independent copy of neural network weights and they need to be trained separately. Therefore, their computational and memory costs increase linearly with the ensemble size in both training and testing~\citep{wen2019batchensemble}. In contrast, although single forward-pass methods have shown to be efficient in modelling uncertainties, they suffered from the so-called feature collapse problem, which can lead to its failure in estimating uncertainties. Specifically, feature collapse can result in the mapping of the features of out-of-distribution (OOD) inputs into the same region of in-distribution sample features~\citep{mukhoti2021deep}.
	
	To take the best from both worlds, we propose the batch-ensemble stochastic neural networks (BE-SNNs), an OOD detection method consisting of a novel single forward-pass approach proposed in this paper and an efficient ensembling mechanism inspired by \citep{wen2019batchensemble}. We assume that different ensemble members of deterministic neural networks can converge to different local minima, thus they will capture different modes of data. Therefore, we can rely on the disagreement between ensemble members to prevent feature collapse by taking predictions given by all members into account. 
	
	Our main contributions are as follows: \newline
	(i) The proposed BE-SNNs can overcome the undesirable feature collapse in single forward-pass methods while also maintaining their low computational cost. We achieve this by constructing an ensemble of single forward pass models with the efficient batch-ensemble mechanism~\citep{wen2019batchensemble}. \newline
	(ii) The novel single forward pass model used in BE-SNNs provides a flexible way to capture the distribution of data features, by representing the distribution of class features with a set of feature vectors produced by a class-dependent feature generator. Within this framework, the empirical distribution may be of more complex structure than simple distributions such as Gaussian with diagonal covariance matrices.\newline
	(iii) We demonstrate the effectiveness of the BE-SNNs on several OOD detection benchmarks, including the Two-Moons dataset, the FashionMNIST vs MNIST dataset, FashionMNIST vs NotMNIST,and the CIFAR-10 vs SVHN dataset.
	
	\section{Preliminaries}
	In this section, we first describe the problem of out-of-distribution (OOD) detection for classification tasks. We then provide a brief introduction to the batch-ensemble mechanism~\citep{wen2019batchensemble}.
	
	\subsection{Problem Statement}
	Denote by $\mathcal{X}$ the input space of a classifier $f:\mathcal{X}\rightarrow\mathbb{R}^{C}$, where $C$ is the number of classes and the output $f(\textbf{x})$ of the classifier predicts the probability of an input sample $\textbf{x}\in\mathcal{X}$ belonging to each class. Let the classifier $f(\cdot)$ be a neural network trained on a dataset drawn from $\mathcal{D}_{\text{in}}$ defined on the input space $\mathcal{X}$, and we denote by $\mathcal{D}_{\text{in}}$ the in-distribution of the classifier $f(\cdot)$. In OOD detection problems, the input sample $\textbf{x}$ during testing time can be drawn from a mixture distribution $\mathcal{D}_{\text{mix}}$ consisting of the in-distribution $\mathcal{D}_{\text{in}}$ and an out-distribution $\mathcal{D}_{\text{out}}$, which is also defined on the input space $\mathcal{X}$. The out-distribution $\mathcal{D}_{\text{out}}$ is supposed to be a data distribution distinct from the in-distribution $\mathcal{D}_{\text{in}}$.
	
	Within this setup, given an input sample $\textbf{x}\sim\mathcal{D}_{\text{mix}}$ drawn from the mixture distribution, the goal of the OOD detection task is to decide whether the input $\textbf{x}$ is an in-distribution sample $\textbf{x}_{\text{in}}\sim\mathcal{D}_{\text{in}}$ or an out-distribution sample $\textbf{x}_{\text{out}}\sim\mathcal{D}_{\text{out}}$. In other words, OOD detectors aim to find a decision function $\mathcal{S}(\textbf{x})$ such that:
	\begin{equation}
		\mathcal{S}(\mathbf{x})= \begin{cases}0 & \text { if }\;\;\textbf{x}\sim\mathcal{D}_{\text{out}} \\ 1 & \text { if }\;\;\textbf{x}\sim\mathcal{D}_{\text{in}}\end{cases}\,,
	\end{equation}
	where $\textbf{x}$ is an input sample drawn from the mixture distribution $\mathcal{D}_{\text{mix}}$. 
	
	To achieve the above goal, one common solution is to compute a specified OOD score of a given input $\textbf{x}$, and classify $\textbf{x}$ as in-distribution if its OOD score is within a given threshold or as an OOD sample otherwise. It is also worth noting that OOD scores can be considered as examples of score functions in conformal outlier detection introduced in Section 6.3 of~\citep{angelopoulos2021gentle}, where the OOD (conformal) threshold is determined by calculating a quantile of inliers' conformal score. However, to comprehensively assess OOD detectors' ability to identify OOD samples, it is required to evaluate their overall performance by taking all different levels of OOD (conformal) thresholds into account. 
	In this work, we compute the OOD score through the proposed batch-ensemble stochastic neural network framework, and evaluate the performance of the proposed method using commonly adopted evaluation metrics such as FPR, AUROC, AUPRC following~\citep{liu2020energy,sun2021react,van2020uncertainty,ren2019likelihood}.
	
	
	\subsection{Batch-ensemble}
	\label{sec:be}
	Inspired by the fact that deep neural networks trained from random initialization can converge to different local optima and thus may not make the same error given the same input, \citep{lakshminarayanan2017simple} proposed deep ensembles. By utilizing a collection of predictions given by different ensemble members, deep ensemble was shown to achieve better performance than an individual neural network and give reliable predictive uncertainty estimates. However, running multiple copies of neural networks both in training and testing hinders further applications of deep ensembles due to high computational costs.
	
	As an alternative, \citep{wen2019batchensemble} proposed batch-ensemble (BE), which is a parameter efficient variant of deep ensemble that construct ensembles over a rank-1 subspace of networks' weights. Denote by $W\in\mathbb{R}^{m\times n}$ the weight matrix of a neural network layer, where $m$ is the input dimension and $n$ the output dimension. Each ensemble member is assigned with a pair of trainable vectors $r_{n_e}\in\mathbb{R}^m$ and $s_{n_e}\in\mathbb{R}^n$, where ${n_e}\in\{1, \cdots, N_e\}$ and $N_e$ is the ensemble size. The weights $\overline{W}_{n_e}$ of ensemble members are then obtained by calculating:
	\begin{align}
		\label{eq:ensemble_weights}
		\overline{W}_{{n_e}}=W \circ F_{{n_e}}\,\,,
	\end{align}
	where $\circ$ refers to the element-wise product, $W$ is shared across ensemble members, and $F_{{n_e}}=r_{{n_e}} s_{{n_e}}^{\top}\in\mathbb{R}^{m\times n}$ is a rank-one matrix which generates the weight matrix for the ${n_e}$-th ensemble member. Given a batch-ensemble layer with input $\textbf{x}$, the output $\textbf{y}$ of the layer can be computed by:
	\begin{align}
		\textbf{y} &=\phi\left(\overline{W}_{{n_e}}^{\top} \textbf{x}\right)=\phi\left(\left(W \circ r_{{n_e}} s_{{n_e}}^{\top}\right)^{\top} \textbf{x}\right) \\
		&=\phi\left(\left(W^{\top}\left(\textbf{x} \circ r_{{n_e}}\right)\right) \circ s_{{n_e}}\right)\,\,,
		\label{eq:be_single}
	\end{align}
	where $\phi$ denotes the activation function. More generally, given a mini-batch $X\in\mathbb{R}^{B\times m}$ with $B$ samples, Equation~(\ref{eq:be_single}) can be vectorized as:
	\begin{align}
		Y=\phi(((X \circ R) W) \circ S)\,\,,
	\end{align}
	where $Y$ and $X$ are mini-batch output and input, respectively, and $R\in\mathbb{R}^{B\times m}$ and $S\in\mathbb{R}^{B\times n}$ are matrices whose rows consist of vectors $r_{n_e}$ and $s_{n_e}$.
	
	With the rank-1 ensemble mechanism described above, the batch-ensemble brings almost no additional computational cost as the only memory overhead is the set of vectors \{$r_1, r_2, \cdots, r_{N_e}$\} and \{$s_1, s_2, \cdots, s_{N_e}$\}, which are much cheaper compared to the full weight matrices. Notably, it is reported in~\citep{wen2019batchensemble} that batch-ensemble of ResNet-32 of size 4 incurs 10\% more parameters, while the vanilla ensemble method incurs 300\% more parameters.
	
	\section{Batch-ensemble Stochastic Neural Networks}
	We present in this section the details of the proposed batch-ensemble stochastic neural networks (BE-SNNs) within a multi-class classification problem. In the following contents. we use BE-SNN-1 to denote a \emph{single} ensemble member of BE-SNNs.
	
	\textbf{Framework of BE-SNNs.}
	For simplicity and without loss of generality, we first introduce how an individual realisation of BE-SNNs, i.e., BE-SNN-1, works. The BE-SNN-1 consists of a data feature extractor $\Phi_\theta: \mathcal{X}\rightarrow\mathbb{R}^d$ mapping the input sample $\textbf{x}\in\mathcal{X}$ to a feature space, and a class feature generator $\Psi_\theta(\cdot)$ outputting the distribution of feature representations of each class. 
	The distribution of class features in the BE-SNN-1, as well as BE-SNNs in general, is represented by a set of feature vectors for each class. This construction enables BE-SNNs to approximate more complex distributions than pre-defined simple distributions such as diagonal Gaussian distributions.
	
	In particular, the class feature generator $\Psi_\theta:\mathbb{R}^{C+d_\epsilon}\rightarrow\mathbb{R}^d$ takes a one-hot class label $\textbf{h}_c\in\mathbb{R}^C$ and a random vector $\epsilon_m\in\mathbb{R}^{d_\epsilon}$ as its input, then outputs $d$-dimensional class-dependent feature vectors $\textbf{e}_{c,m}$:
	\begin{equation}
		\textbf{e}_{c,m}=\Psi_\theta(\textbf{h}_c, \epsilon_m)\;\; \text{for}\;\;\forall m\in\{1, \cdots, M\};\;\forall c\in\{1, \cdots, C\}\,\,,
	\end{equation}
	where $\epsilon_m{\sim}\mathcal{N}(\textbf{0},\mathbb{I}_{d_\epsilon})$, $M$ is the number of feature vectors for each class, and the distribution of the $c$-th class in the feature space is represented by a collection of feature vectors $\textbf{E}_{c}:=\{\textbf{e}_{c,m}\}_{m=1}^{M}$.
	
	The BE-SNN-1 classifies a given input sample $\textbf{x}$ based on the distance from extracted input features $\Phi_\theta(\textbf{x})$ to the feature representation $\textbf{E}_{c}$ of each class. Here, we adopt the average squared Euclidean distance $\bar{\textbf{d}}_c(\textbf{x})=\frac{1}{M}\sum_{m=1}^{M}||\Phi_\theta(\textbf{x})-\textbf{e}_{c,m}||_2^2$ as the distance metric, where $||\cdot||_2$ denotes the $L_2$-norm. Each time BE-SNN-1 makes a prediction, the class feature extractor will generate a set of feature vectors $\textbf{E}_{c}:=\{\textbf{e}_{c,m}\}_{m=1}^{M}$ by drawing $M$ random samples $\{\epsilon_m\}_{m=1}^M$ from the standard Gaussian distribution $\mathcal{N}(\textbf{0},\mathbb{I}_{d_\epsilon})$, and classify the input $\textbf{x}$ as the class with the minimum distance to the feature vector of the input, which is equivalent to the class with the maximum kernel value:
	\begin{equation}
		\label{eq:snn_class}
		\underset{c}{\operatorname{argmax}}(\exp(-\bar{\textbf{d}}_c(\textbf{x})))=\underset{c}{\operatorname{argmax}}\big\{\exp(-\frac{1}{M}\sum_{m=1}^{M}||\Phi_\theta(\textbf{x})-\textbf{e}_{c,m}||_2^2)\big\}\,\,.
	\end{equation}
	
	However, as discussed in Section~\ref{sec:introduction}, evaluating the uncertainty by modelling the distribution of data features may suffer from feature collapse. Therefore, we now discuss how to utilize the batch-ensemble mechanism to avoid feature collapse and introduce the proposed batch-ensemble stochastic neural networks (BE-SNNs).
	
	Firstly, we build an ensemble of BE-SNN-1s by using batch-ensemble layers introduced in Section~\ref{sec:be}, thus each batch-ensemble layer has their own trainable vectors $\{r_{l,n_e}\}_{n_e=1}^{N_e}$ and $\{s_{l,n_e}\}_{n_e=1}^{N_e}$, where $l$ is the layer index, and $N_e$ is the ensemble size. Since the weights defined in Equation~(\ref{eq:ensemble_weights}) are dependent on the trainable vectors $\{r_{l,n_e}\}_{n_e=1}^{N_e}$ and $\{s_{l,n_e}\}_{n_e=1}^{N_e}$, the output layer of BE-SNNs has $N_e$ different outputs.
	As a result, the data feature extractors $\{\Phi_{\theta_{n_e}}(\cdot)\}_{n_e=1}^{N_e}$ in BE-SNNs produce $N_e$ different feature vectors for a given input $\textbf{x}$. And the class feature generators $\{\Psi_{\theta_{n_e}}(\cdot)\}_{n_e=1}^{N_e}$  produce $N_e\times M$ feature vectors for each class.
	
	To make a prediction for a given input $\textbf{x}$, the BE-SNNs classify $\textbf{x}$ as the class with the maximum average kernel value:
	\begin{equation}
		\underset{c}{\operatorname{argmax}}\big\{\frac{1}{N_e}\sum_{n_e=1}^{N_e}\exp(-\frac{1}{M}\sum_{m=1}^{M}||\Phi_{\theta_{n_e}}(\textbf{x})-\textbf{e}_{n_e,c,m}||_2^2)\big\}\,\,.
	\end{equation}
	\textbf{Loss functions of BE-SNNs.}
	We first introduce the loss function for an individual ensemble member of BE-SNNs. The loss function of BE-SNN-1 consists of two parts, a classification loss and regularization terms. For the classification loss, we use the cross entropy loss between the exponentiated negative distance $\tilde{\textbf{y}}_c=\exp(-\bar{\textbf{d}}_c(\textbf{x}))$ and the one-hot ground-truth class label $\textbf{y}$:
	\begin{equation}
		L_{\text{cls}}(\tilde{\textbf{y}}, \textbf{y})=-\sum_{c=1}^{C}\textbf{y}_c\log(\tilde{\textbf{y}}_c)+(1-\textbf{y}_c)\log(1-\tilde{\textbf{y}}_c)\,\,,
	\end{equation}
	%
	%
	where $\textbf{y}_c$ indicates the $c$-th element of the one-hot ground-truth label $\textbf{y}$. 
	
	\textbf{Entropy-based Regularization.}
	To prevent class feature vectors $\textbf{E}_{c}:=\{\textbf{e}_{c,m}\}_{m=1}^{M}$ from degenerating to single point estimates, i.e. feature vectors $\textbf{e}_{c,m}$ converging to almost the same value for all $m\in\{1,\cdots, M\}$, we add a regularization term ${R}_c(\cdot)$ into the loss function.
	Particularly, we design the regularization term ${R}_c(\cdot)$ to encourage the entropy of the $c$-th class feature to be proportional to that of the data sample feature belonging to the $c$-th class. To achieve this, we first approximate the entropy $\hat{\mathbb{H}}(\{\Phi_\theta(\textbf{x})|\textbf{y}_c=1\})$ of the data sample feature for all $c\in\{1, \cdots, C\}$ using the $k$-nearest neighbour entropy estimator proposed in~\citep{lombardi2016nonparametric}. The class-dependent entropy $\hat{\mathbb{H}}(\{\Phi_\theta(\textbf{x})|\textbf{y}_c=1\})$ is then used as a threshold to penalise class features whose entropy is smaller than this threshold. The described regularization term can be computed as follows:
	\begin{align}
		{R}_c(\textbf{E}_{c})&={R}_c(\{\textbf{e}_{c,m}\}_{m=1}^M)\\
		&=\operatorname{max}\bigg(0,\; \hat{\mathbb{H}}(\{\Phi_\theta(\textbf{x})|\textbf{y}_c=1\})- \hat{\mathbb{H}}(\{\textbf{e}_{c,m}\}_{m=1}^M)\bigg)\,\,.
	\end{align}
	In addition, the loss function of BE-SNN-1 also includes a gradient penalty term $L_{\text{gp}}(\tilde{\textbf{y}})$ to encourage sensitivity of the classifier as in~\citep{van2020uncertainty}:
	\begin{equation}
		L_{\text{gp}}(\tilde{\textbf{y}})=\left[\left\|\nabla_{\textbf{x}} \sum_{c} \tilde{\textbf{y}}_{c}\right\|_{2}^{2}-1\right]^{2}\,\,,
	\end{equation}
	where $\nabla_{\textbf{x}} \sum_{c} \tilde{\textbf{y}}_{c}$ is the derivative of $\sum_{c} \tilde{\textbf{y}}_{c}$ w.r.t to the input $\textbf{x}$. The overall loss function for the $n_e$-th ensemble member of the BE-SNN model is as follows:
	\begin{equation}
		\label{eq:loss_snn}
		L_{n_e}=L_{\text{cls}}(\tilde{\textbf{y}}, \textbf{y}) +\lambda_1 \sum_{c=1}^{C}{R}_c(\textbf{E}_{c}) + \lambda_2 L_{\text{gp}}(\tilde{\textbf{y}})\,\,.
	\end{equation}
	
	To train BE-SNNs, we minimize the average loss functions across all ensemble members, the loss function of BE-SNNs is:
	\begin{equation}
		L_{\text{BE-SNNs}}=\frac{1}{N_e}\sum_{n_e=1}^{N_e}L_{ n_e}\,\,.
	\end{equation}
	
	\textbf{OOD detection score.} 
	Since softmax activations can produce arbitrarily high confidence even for OOD samples~\citep{liu2020energy}, we propose to adopt an adaptively tempered variant of softmax activation in the BE-SNNs for detecting OOD samples. Notably, this modified version of softmax is only used for OOD detection, but not for training and making predictions for in-distribution samples.
	
	Given a BE-SNN-1, denote by $\bar{\textbf{d}}(\textbf{x})=[\bar{\textbf{d}}_1(\textbf{x}),\cdots \bar{\textbf{d}}_C(\textbf{x})]$ the distance vector for each class, we compute a categorical distribution by using the tempered negative distance
	\begin{equation}
		\xi(\textbf{x})=\frac{-\bar{\textbf{d}}(\textbf{x})}{\exp(\underset{c}{\operatorname{min}}(\bar{\textbf{d}}_c(\textbf{x})))}
		\label{eq:temperd_neg_dist}
	\end{equation}
	as the input of a standard softmax function $\sigma(\cdot)$, where $\underset{c}{\operatorname{min}}(\bar{\textbf{d}}_c(\textbf{x}))$ is the smallest distance between class features and input features. It is worth mentioning that $\sigma(\xi(\textbf{x}))$ can be seen as a tempered softmax prediction as in~\citep{guo2017calibration}, whereas the tempering constant $\exp(\underset{c}{\operatorname{min}}(\bar{\textbf{d}}_c(\textbf{x})))$ is data-adaptive. 
	
	An intuitive interpretation of the above modification is that the further an input feature is from class features, the more uniform the categorical distribution should be. This is realized by the introduction of the adaptive tempering constant $\exp(\underset{c}{\operatorname{min}}(\bar{\textbf{d}}_c(\textbf{x})))$ to avoid producing arbitrarily high confidence on OOD samples.
	
	We propose to use the entropy of the tempered softmax prediction $\mathbb{H}[{\zeta}(\textbf{x})]=-\sum_{c}{\bar{\omega}}_{c}\log{\bar{\omega}}_{c}$ as the OOD score, where
	\begin{gather}
		{\zeta}(\textbf{x})=[\bar{\omega}_{1}, \cdots, \bar{\omega}_{C}]\,\,,\\
		\bar{\omega}_{c}=\frac{1}{N_e}\sum_{n_e=1}^{N_e}\sigma_c(\xi_{n_e}(\textbf{x}))\,\,,
	\end{gather}
	and $\xi_{n_e}(\textbf{x})$ is the tempered negative distance of the $n_e$-th ensemble member computed in Equation~(\ref{eq:temperd_neg_dist}). The decision function in the BE-SNNs is as follows:
	\begin{equation}
		\mathcal{S}(\mathbf{x})= \begin{cases}\text{out-of-distribution} & \text { if }\;\; \mathbb{H}[{\zeta}(\textbf{x})]\geq \tau \\ \text{in-distribution} & \text { if }\;\;\mathbb{H}[{\zeta}(\textbf{x})]< \tau\end{cases}\,,
	\end{equation}
	where $\tau$ is a user-specified threshold.

	\section{Experiments}
	
	In this section, we evaluate the performance of BE-SNNs on several OOD benchmarks adopted by previous works~\citep{van2020uncertainty,wan2018rethinking}, including the Two-Moons dataset, the FashionMNIST vs MNIST dataset, the FashionMNIST vs NotMNIST dataset, and the CIFAR10 vs SVHN dataset. In all the experiments, we set the regularization coefficient $\lambda_1$ in Equation~(\ref{eq:loss_snn}) to be $1.0$, and set $\lambda_2$ to be $0.5$. Details on network architectures employed in the experiment can be found in Appendix~\ref{appendix:architecture}.
	\subsection{Two Moons dataset}
	\begin{figure*}[hbt!]
		\begin{center}
			\begin{tabular}{ccc}
				\includegraphics[height=35mm]{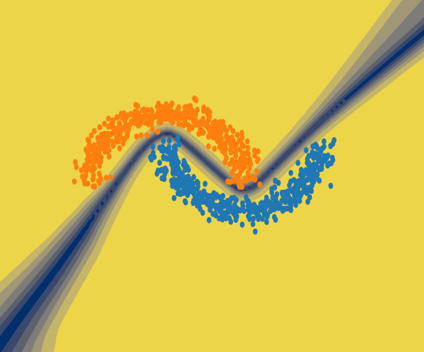} &  \includegraphics[height=35mm]{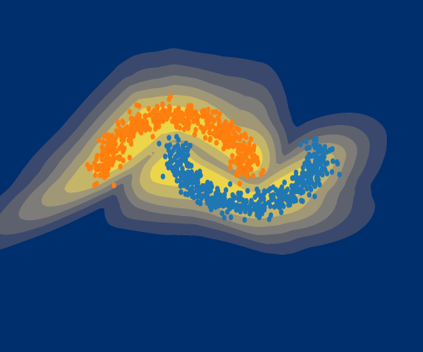} & \includegraphics[height=35mm]{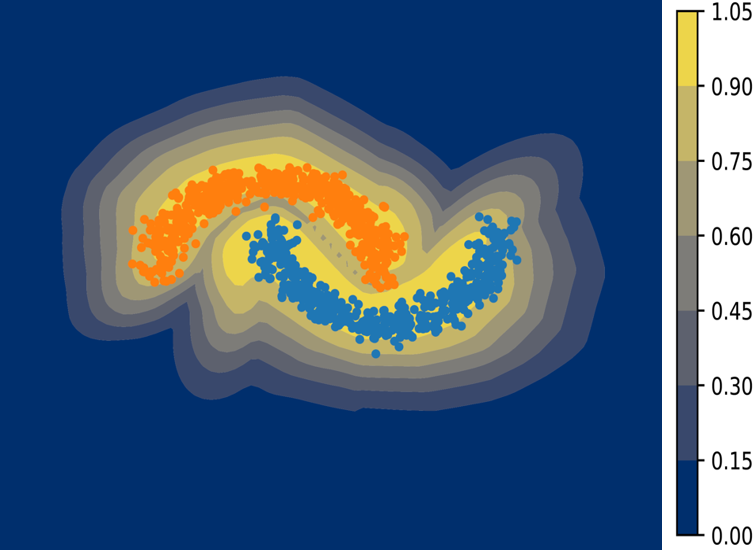}\\
				
				(a) Deep ensemble & (b) DUQ &(c) BE-SNN-1 \\
			\end{tabular}
			\caption{\small Visualized experimental results of deep ensemble, DUQ, and BE-SNN-1 on the Two-Moons dataset. The confidence of models in their predictions is indicated by different colours, where blue corresponds to low confidence, and yellow shows high confidence. Left plot shows that deep ensemble only has low confidence on regions around its decision boundary. In contrast, BE-SNN-1 only produce high confidence predictions on the regions where the model was trained on, and show gradually decreasing confidence on the other areas far away from the training data. DUQ produces slightly more mismatch between the high confidence regions and the samples which the model was trained with compared to the BE-SNN-1.}
			\label{fig:two_moons}
		\end{center}
	\end{figure*}
	Following the setup in~\citep{van2020uncertainty}, we first visualize the performance of the BE-SNN-1 on the Two-Moons dataset and compare it with the DUQ and a deep ensemble of softmax networks with ensemble size of 4. As shown in Figure~\ref{fig:two_moons}, the Two-Moons dataset is a 2-dimensional classification dataset. In this experiment, while the evaluated dataset is a toy dataset, we do not only focus on the classification accuracy since both models can achieve 100\% accuracy, but also how well the tested models can assign proper confidence scores to their predictions. Notably, for the DUQ and BE-SNN-1, we consider the exponentiated distance to the closet centroid as the confidence of the prediction, and the confidence of deep ensemble methods is evaluated by the maximum value of the averaged softmax prediction.
	
	The experiment results presented in Figure~\ref{fig:two_moons} indicate that, even with deep ensembles of softmax network, it can still produce overconfident predictions on OOD samples. A possible reason is that the Two-Moons dataset is too simple for the networks to converge to different local optima, thus all ensemble members are almost the same after training. As a comparison, both the BE-SNN-1 and the DUQ are able to assign proper confidence to their predictions, since it can be observed from Figure~\ref{fig:two_moons} that they only produce high confidence predictions on the in-distribution, and show gradually decreasing confidence on the other areas far away from the in-distribution.
	
	\subsection{Image classification datasets}
	\begin{table}[b]
		\centering
		\resizebox{0.9\textwidth}{!}{
			\begin{tabular}{|c|c|c|c|c|c|c|}
				\hline
				In-Distibution                & Model      & OOD & Accuracy$\uparrow$  &  FPR95$\downarrow$ & AUROC$\uparrow$ & AUPRC$\uparrow$ \\ \hline
				\multirow{7}{*}{FashionMNIST} & \multirow{2}{*}{BE-SNNs}    & MNIST& \multirow{2}{*}{\textbf{92.4}\textbf{\%}$\boldsymbol{\pm}$\textbf{0.1}\%} &   \textbf{0.169}$\pm$\textbf{.016}   &  \textbf{0.961}\boldsymbol{$\pm$}\textbf{.016}    & \textbf{0.989}\boldsymbol{$\pm$}\textbf{.002}\\\cline{3-3} \cline{5-7} 
				&&NotMNIST&&\textbf{0.182}\boldsymbol{$\pm$}\textbf{.057}&\textbf{0.970}\boldsymbol{$\pm$}\textbf{.018}& 0.988$\pm$.005\\\cline{2-7}
				& \multirow{2}{*}{DUQ}                 &  MNIST   &   \multirow{2}{*}{92.1\%$\pm$0.1\%}  &    0.244$\pm$0.043   &  0.944$\pm$0.013  &  0.975$\pm$.004 \\\cline{3-3} \cline{5-7} 
				&&NotMNIST&&0.231$\pm$0.047 &0.953$\pm$0.009&\textbf{0.989}\boldsymbol{$\pm$}\textbf{.007}\\\cline{2-7}
				& \multirow{2}{*}{Gaussian}                 &  MNIST &\multirow{2}{*}{92.0\%$\pm$0.1\%}   &    0.274$\pm$0.041   &  0.932$\pm$0.014  & 0.980$\pm$.008  \\\cline{3-3} \cline{5-7} 
				&&NotMNIST&&0.281$\pm$0.061 &0.940$\pm$0.012&0.982$\pm$.003\\\cline{2-7} 
				& \multirow{2}{*}{Softmax}                 &  MNIST &\multirow{2}{*}{92.3\%$\pm$0.2\%}   &    0.564$\pm$0.051   &  0.889$\pm$0.034  & 0.930$\pm$.015 \\\cline{3-3} \cline{5-7} 
				&&NotMNIST&&0.531$\pm$0.121 &0.910$\pm$0.013&0.941$\pm$.009\\\hline
				
				\multirow{3}{*}{CIFAR-10}     & BE-SNNs                &   SVHN  &    93.5\%$\pm$ 0.1\%  & \textbf{0.359}\boldsymbol{$\pm$}\textbf{.031} &  \textbf{0.940}\boldsymbol{$\pm$}\textbf{0.017}   & \textbf{ 0.982}\boldsymbol{$\pm$}\textbf{.007}\\ \cline{2-7} 
				& DUQ                 &  SVHN   &    93.7\%$\pm$ 0.2\%   & 0.452$\pm$.035 &  0.921$\pm$0.013  &  0.970$\pm$.006\\ \cline{2-7} 
				& Gaussian &  SVHN   &    93.1\%$\pm$ 0.1\%  &  0.492$\pm$.045  & 0.919$\pm$0.009 & 0.975$\pm$.003\\ \cline{2-7} 
				& Softmax &  SVHN   &    \textbf{93.9}\%$\pm$ \textbf{0.1}\%  &  0.680$\pm$.032  & 0.872$\pm$0.024 & 0.924$\pm$.011\\ \hline
			\end{tabular}%
		}
		\caption{\small Experiments results on FashionMNIST vs MNIST, FashionMNIST vs NotMNIST, and CIFAR-10 vs SVHN datasets. Compared baselines are the DUQ~\citep{van2020uncertainty}, and the Gaussian classifier~\citep{wan2018rethinking}. The lower the FPR95, the higher the AUROC and the AUPRC, the better the performance on the OOD detection task. The mean and standard deviation are calculated over 5 random seeds.}
		\label{tab:experiment_comparison}
	\end{table}
	In this section, we evaluate the BE-SNNs' ability to detect OOD samples on several image OOD detection datasets, including the FashionMNIST vs MNIST dataset, the FashionMNIST vs NotMNIST dataset, and the CIFAR10 vs SVHN dataset. Three baselines are compared with the BE-SNNs, including a vanilla softmax neural network, the DUQ~\citep{van2020uncertainty}, and the Gaussian classifier~\citep{wan2018rethinking}. While the DUQ and the Gaussian classifier adopt the negative likelihood of the predicted class as the OOD score, the predictive entropy is used in the softmax neural network. 
	
	As for the evaluation metrics, we report following three OOD detection metrics in the experiment results as in previous works~\citep{liu2020energy,kim2021locally,sun2021react}: (i) FPR95, the false positive rate of classifying OOD examples when the true positive rate (recall) of in-distribution is 95\%; (ii) the area under the ROC curve (AUROC); (iii) the area under the precision-recall curve (AUPRC). Lower FPR95, higher AUROC and AUPRC indicate better performance in OOD detection tasks.
	
	\textbf{FashionMNIST vs MNIST \& NotMNIST dataset.} In this experiment, we train evaluated models on the FashionMNIST dataset, and expect the models to be able to distinguish FashionMNIST (in-distribution) samples from MNIST and NotMNIST (out-of-distribution) samples based on their OOD scores. The feature extractors of DUQ, Gaussian classifier, and softmax networks are a convolutional network consisting of three convolutional layers followed by a fully-connected output layer, and the BE-SNNs are constructed by the batch-ensemble variants of the same convolutional network. The class feature generator of BE-SNNs is simply a two layer fully-connected batch-ensemble network, and the softmax network uses a fully-connected layer with softmax activation function to classify the output of the feature extractor. The experiment results of BE-SNNs shown in Table~\ref{tab:experiment_comparison} are achieved by using $N_e=4$ ensemble members. To draw a fair comparison, we design the network architectures of evaluated methods to have similar number of parameters to be optimized as shown in Table~\ref{tab:runtime}. 
	
	It can be observed from Table~\ref{tab:experiment_comparison} that all the evaluated methods produced similar classification accuracy, while the evaluated OOD detection metrics varied among different methods. The softmax network leads to the worst OOD detection metrics on both MNIST and NotMNIST datasets. The BE-SNNs achieved comparable or even better OOD detection performance in most settings than the evaluated baselines in terms of the FPR95, AUROC, and AUPRC. Regarding the runtime, the softmax network is the most efficient approach as expected. The DUQ and the Gaussian classifier have similar computational costs, while the BE-SNNs with 4 ensemble members require about 50\% more time to complete a forward propagation for a mini-batch containing 500 image samples. The relatively high computational costs in single-forward pass methods are mainly caused by the computation of distance matrices of input features and class features, which is more time-consuming than a simple fully-connected layer in softmax networks.
	
	We also compared the performance of BE-SNNs with different number of ensemble members in Table~\ref{tab:different_ensemble_sizes}. The experiment results in Table~\ref{tab:different_ensemble_sizes} demonstrate that as the ensemble size $N_e$ increases, the BE-SNNs produce better performance considering the classification accuracy and OOD detection metrics. However, the improvement from $N_e=4$ to $N_e=8$ is only marginal, this is possibly due to training method of batch-ensemble layers. Particularly, when training BE-SNNs, we repeat a mini-batch with $B$ samples $N_e$ times so that each ensemble member receives the whole batch of samples. We then keep the product $B\times N_e$ a constant as we change $N_e$, which implies that increasing $N_e$ will lead to a smaller batch size. This provides one plausible reason that the decreasing batch size for $N_e=8$ did not lead to better performance than $N_e=4$ that require further investigation. In addition, we also conducted ablation studies to investigate the effect of the adaptively tempered softmax, and the gradient penalty regularization on the performance of BE-SNNs. Details of ablation studies can be found in Appendix~\ref{appendix:ablation}.
	
	\begin{table}[t]
		\centering
		\begin{tabular}{|c|c|c|}
			\hline
			Methods  & \# Parameters & Runtime per batch\\\hline
			BE-SNNs  & 1136878    & 5.9 $\times 10^{-3}$/s  \\ \hline
			DUQ      & 1009408    & 3.9 $\times 10^{-3}$/s  \\ \hline
			Gaussian & 1080320    & 3.7 $\times 10^{-3}$/s  \\ \hline
			Softmax  & 1152778    & 1.4 $\times 10^{-3}$/s  \\ \hline
		\end{tabular}
		\caption{\small Number of parameters and runtime of the evaluated methods. The runtime refers to the computational time of one forward propagation for a mini-batch containing 500 samples, and is calculated based on a computer with an Intel(R) Core(TM) i9 @2.50GHz, 2496MHz 8 core processor, and a RTX 3090 graphic card with 64GB RAM and 24GB GPU memory.}
		\label{tab:runtime}
	\end{table}
	
	\textbf{CIFAR-10 vs SVHN dataset.} We have also evaluated the performance of BE-SNNs on the CIFAR-10 dataset, with the SVHN dataset as OOD dataset. In this experiment, ResNet-18~\citep{he2016deep} followed by an additional fully-connected output layer is used as the feature extractor. From Table~\ref{tab:experiment_comparison}, we observe that the BE-SNNs present comparable classification accuracy to the other two baselines. Notably, the BE-SNNs have significantly lower FPR95, higher AUROC and AUPRC than the DUQ and the Gaussian classifier, indicating that the BE-SNNs have better performance on the OOD detection task than the two baselines.
	
	\begin{table}[h]
		\centering
		\resizebox{0.9\textwidth}{!}{
			\begin{tabular}{|c|c|c|c|c|c|c|}
				\hline
				In-Distibution                & BE-SNNs     & OOD & Accuracy$\uparrow$  &  FPR95$\downarrow$ & AUROC$\uparrow$ & AUPRC$\uparrow$ \\ \hline
				\multirow{8}{*}{FashionMNIST} & \multirow{2}{*}{$N_e=1$}    & MNIST&   \multirow{2}{*}{92.2\%$\pm$0.1\%}  &    0.246$\pm$0.031   &  0.945$\pm$0.006  &  0.983$\pm$.004\\\cline{3-3} \cline{5-7} 
				&&NotMNIST&&0.197$\pm$0.048 &0.963$\pm$0.006&0.987$\pm$.007\\\cline{2-7}	
							
				& \multirow{2}{*}{$N_e=2$}    & MNIST&   \multirow{2}{*}{92.2\%$\pm$0.1\%}  &    0.204$\pm$0.052   &  0.956$\pm$0.011  &  0.985$\pm$.005 \\\cline{3-3} \cline{5-7} 
				&&NotMNIST&&0.211$\pm$0.042 &0.955$\pm$0.013&0.984$\pm$.006\\\cline{2-7}
				
				& \multirow{2}{*}{$N_e=4$}                 &  MNIST   & \multirow{2}{*}{\textbf{92.4}\%$\pm$\textbf{0.1}\%} &   \textbf{0.169}$\pm$\textbf{.016 }  &  0.961$\pm$.016    & 0.989$\pm$.002\\\cline{3-3} \cline{5-7} 
				&&NotMNIST&&0.182$\pm$.057&\textbf{0.970}$\pm$\textbf{.018}& \textbf{0.988}$\pm$\textbf{.005}\\\cline{2-7}
				
				& \multirow{2}{*}{$N_e=8$}                 &  MNIST &\multirow{2}{*}{92.4\%$\pm$0.2\%}   &    0.174$\pm$0.032   &  \textbf{0.972}$\pm$\textbf{0.011}  & \textbf{0.990}$\pm$\textbf{.004}  \\\cline{3-3} \cline{5-7} 
				&&NotMNIST&&\textbf{0.181}$\pm$\textbf{0.047} &0.969$\pm$0.010&0.987$\pm$.004\\\hline
			\end{tabular}%
		}
		\caption{\small Experiments results of BE-SNNs with different ensemble sizes $N_e\in\{2, 4, 8\}$ on the FashionMNIST vs MNIST and FashionMNIST vs NotMNIST datasets.  Increasing the ensemble size $N_e$ will ideally leads to improved performance, we suspect that the decreasing batch size is the reason for the similar performances of $N_e=4$ and $N_e=8$. The mean and standard deviation are calculated over 5 random seeds.}
		\label{tab:different_ensemble_sizes}
	\end{table}	
	
	\section{Conclusion}
	In this work, we proposed the batch-ensemble stochastic neural networks (BE-SNNs), an OOD detection approach that incorporates the batch-ensemble mechanism with a novel single forward pass uncertainty quantification framework. By aggregating the predictions given by different ensemble members, BE-SNNs are algorithmically designed to overcome the feature collapse problem with deterministic single-forward pass models. Besides, BE-SNNs are memory efficient and has low computational cost. We also evaluated the performance of BE-SNNs on several OOD detection benchmarks and compared BE-SNNs with other state-of-the-art OOD detection approaches. Experiment results showed that BE-SNNs have superior performance on both toy and real-world image datasets over the other evaluated methods.
	\bibliographystyle{abbrvnat}
	\bibliography{ref.bib}
	\newpage
	
	\begin{appendices}

		\section{\texorpdfstring{Network Architectures}{Appendices}}
		\label{appendix:architecture}
		In this section, we provide the network architectures employed in the experiment section. We denote a convolutional layer whose kernel size is $s$ with $K$ kernels by $\text{Conv}_{K}(s\times s)$, and a fully-connected layer whose input and output layer have $s_1$ and $s_2$ neurons by $\text{FC}(s_1\times s_2)$. Correspondingly, the batch-ensemble variants of convolutional layer and fully-connected layer are denoted by $\text{BE-Conv}_{C}(s\times s)$ and $\text{BE-FC}(s_1\times s_2)$, respectively. A max pooling layer with kernel size $s\times s$ is denoted by $\text{MaxPooling}(s\times s)$.
		\subsection{Network Architecture in the Two Moons Experiment}
		The network structure of the BE-SNN-1 in the Two-Moons experiment is as follows:
			\begin{align*}
			&\text{Feature extractor}\;\;\Phi_\theta: \\
			&\text{BE-FC}(2\times32)\rightarrow \text{ReLU}\rightarrow \text{BE-FC}(32\times32)\rightarrow \text{ReLU}\rightarrow \text{BE-FC}(32\times32)\\
			&\text{Class feature generator}\;\; \Psi_\theta:\\
			&\text{BE-FC}((2+8)\times 32)\rightarrow \text{ReLU}\rightarrow \text{BE-FC}(32\times32)\\
		\end{align*}
		The input dimension of feature extractor $\Phi_\theta$ equals 2 because the Two-Moons dataset is a 2-dimensional dataset, and the input dimension of class feature generator $\Psi_\theta$ is 10, which equals the number of classes $C=2$ in the Two-Moons dataset plus the noise dimension $d_\epsilon=8$. The outputs of both networks are 32-dimensional feature vectors.
		
		\subsection{Network Architecture in the Image Classification Experiment}
		The network structure of the BE-SNNs in the FashionMNIST vs MNIST \& NotMNIST experiment is as follows:
		\begin{align*}
			&\text{Feature extractor}\;\;\Phi_\theta: \\
			&\text{BE-Conv}_{64}(3\times3)\rightarrow \text{BatchNorm}\rightarrow \text{ReLU}\rightarrow \text{MaxPooling}(2\times2)\rightarrow\\
			& \text{BE-Conv}_{128}(3\times3)\rightarrow \text{BatchNorm}\rightarrow \text{ReLU}\rightarrow \text{MaxPooling}(2\times2)\rightarrow\\
			& \text{BE-Conv}_{128}(3\times3)\rightarrow \text{BatchNorm}\rightarrow \text{ReLU}\rightarrow \text{MaxPooling}(2\times2)\rightarrow\\
			&\text{Flatten}\rightarrow \text{BE-FC}(512\times256)\rightarrow \text{ReLU}\rightarrow \text{BE-FC}(256\times 256)\\
			&\text{Class feature generator}\;\; \Psi_\theta:\\
			&\text{BE-FC}((10+512)\times 512)\rightarrow \text{ReLU}\rightarrow \text{BE-FC}(512\times 256)
		\end{align*}
		The input dimension of class feature generator $\Psi_\theta$ is 522, which equals the number of classes $C=10$ in the FashionMNIST vs MNIST \& NotMNIST dataset plus the noise dimension $d_\epsilon=512$. The outputs of both networks are 256-dimensional feature vectors.
		
		For the CIFAR-10 vs SVHN dataset, we use the batch-ensemble variant of the ResNet-18~\citep{he2016deep} as the backbone of the feature extractor:
		\begin{align*}
			&\text{Feature extractor}\;\;\Phi_\theta: \\
			&\text{BE-ResNet-18}\rightarrow  \text{BE-FC}(512\times 512)\\
			&\text{Class feature generator}\;\; \Psi_\theta:\\
			&\text{BE-FC}((10+512)\times 512)\rightarrow \text{ReLU}\rightarrow \text{BE-FC}(512\times 512)
		\end{align*}
		Note that the other evaluated methods employ similar neural network architectures described above, except that the network width and kernel size are modified to match their number of parameters.
		\section{\texorpdfstring{Ablation Study}{Appendices}}
		In this section we use the FashionMNIST vs MNIST dataset as an example to investigate the effect of the adaptively tempered softmax, the entropy-based regularization, and the gradient penalty regularization on the performance of BE-SNNs. 
		
\begin{figure}[H]
			\begin{center}
				\begin{tabular}{ccc}
					\includegraphics[width=0.3\linewidth]{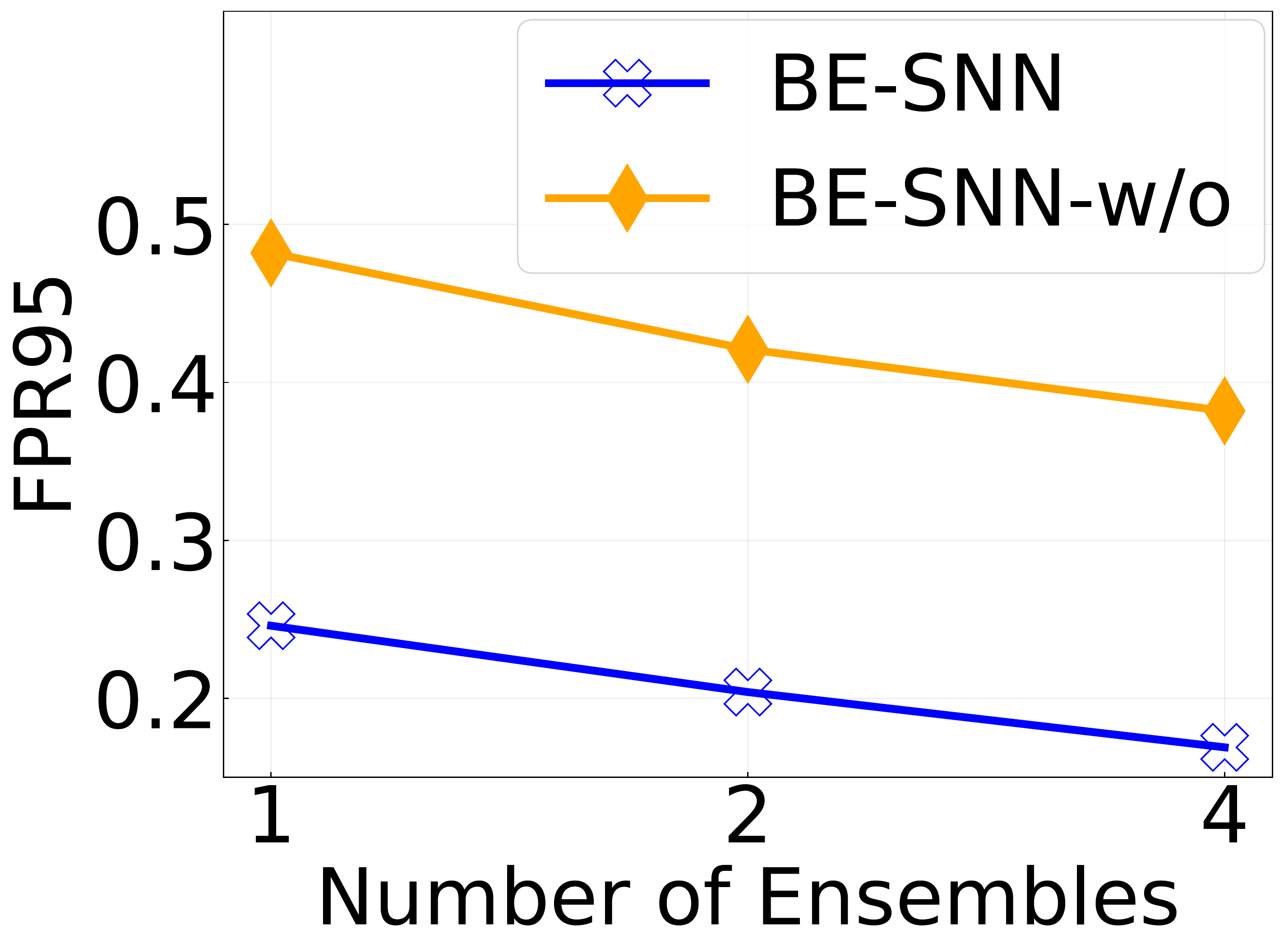}&\includegraphics[width=0.3\linewidth]{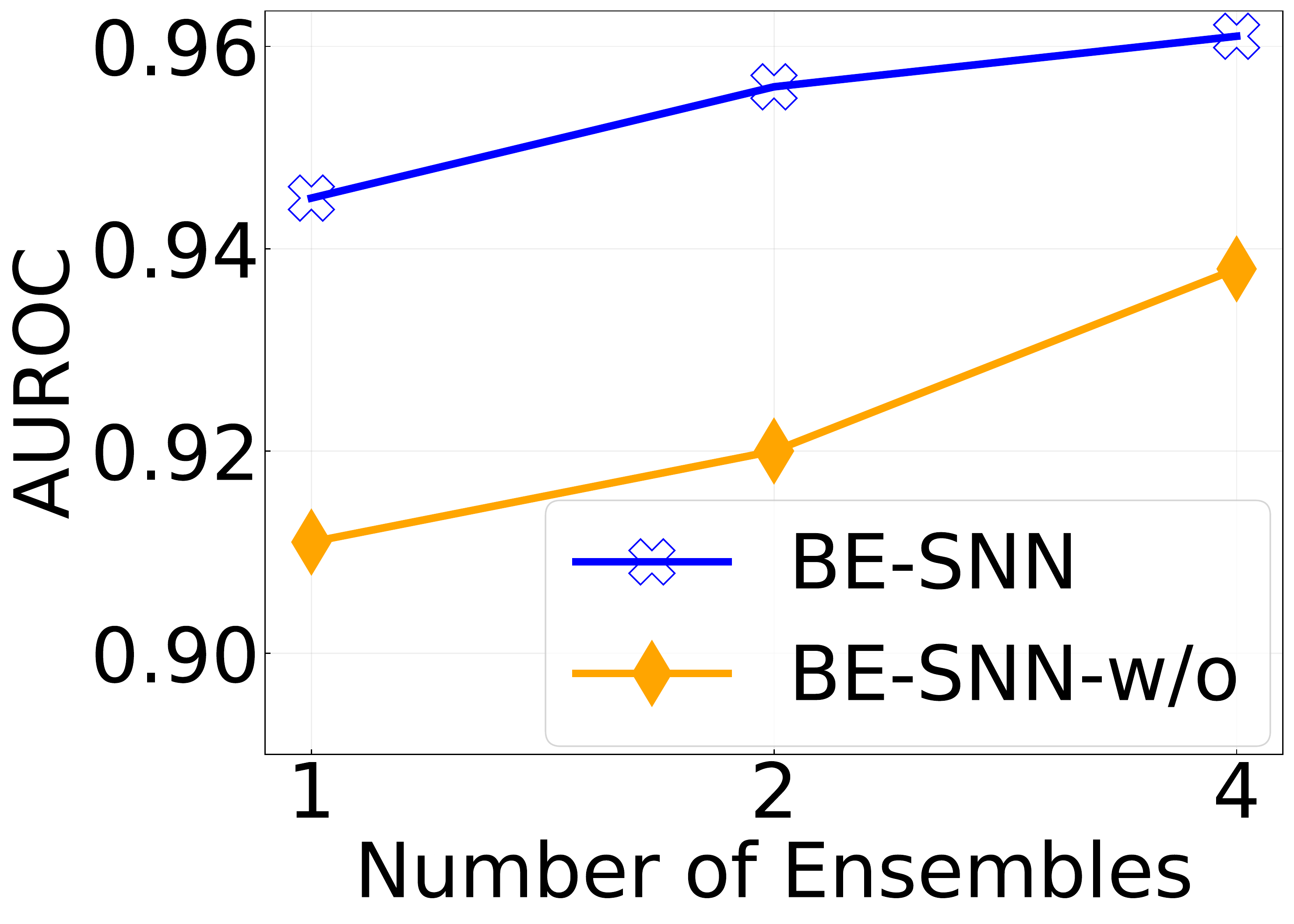} & \includegraphics[width=0.3\linewidth]{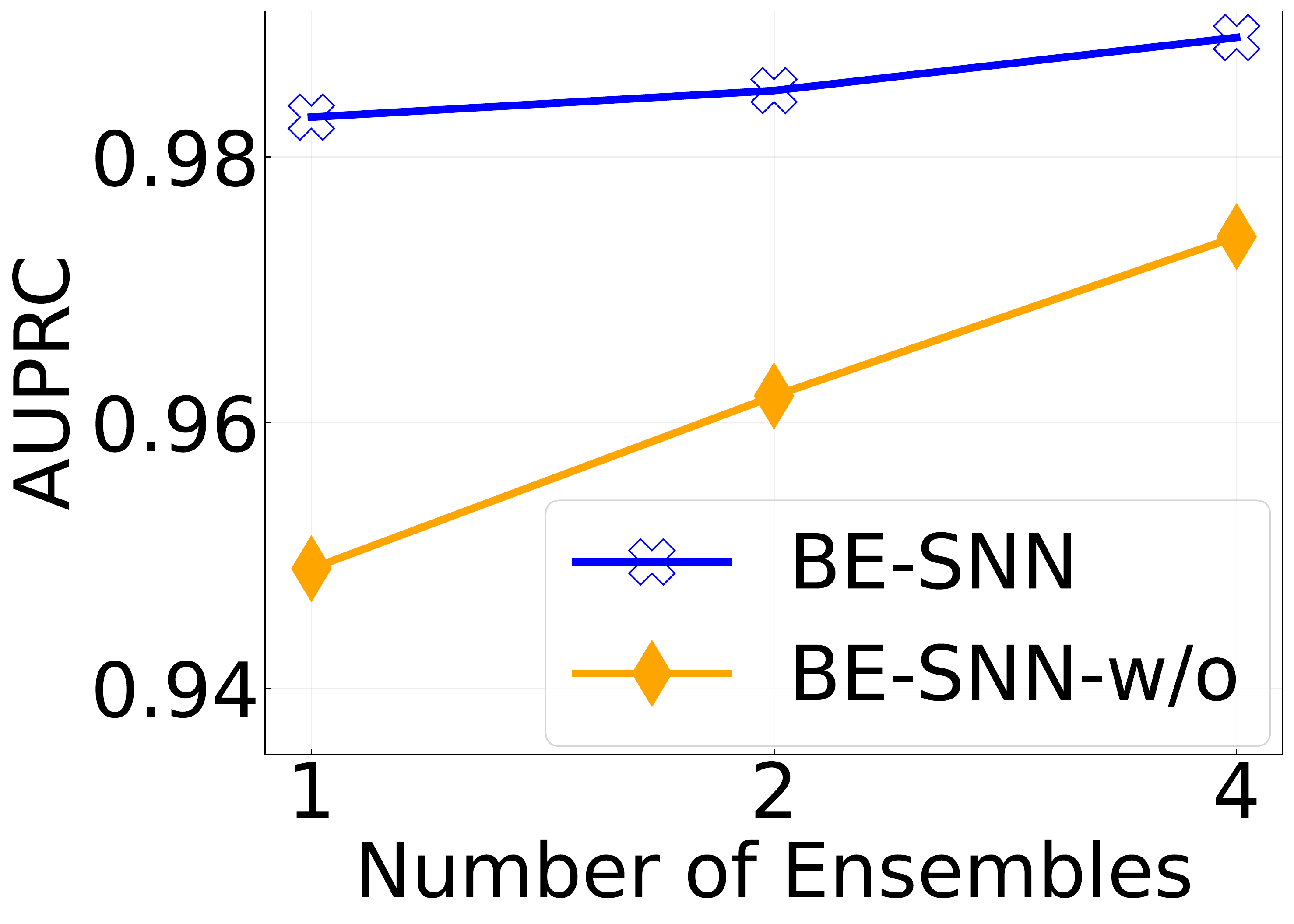}\\
					(a) FPR95 &(b) AUROC & (c) AUPRC\\

				\end{tabular}
				\caption{Comparisons of the OOD detection performance between BE-SNNs with and without adaptively tempered softmax. We use ``BE-SNN-w/o'' to label the BE-SNNs without adaptively tempered softmax. Horizontal axis is the number of ensemble members in BE-SNNs, and the vertical axis is OOD detection metric. Lower FPR95, higher AUROC and AUPRC indicate better performance.}
				\label{figure:adaptive}
			\end{center}
		\end{figure}

		\textbf{Adaptively tempered softmax:} We first evaluate the impact of adaptively tempered softmax by comparing the performance of BE-SNNs with and without the tempered softmax. From Figure~\ref{figure:adaptive}, we can observe that the BE-SNNs equipped with the adaptively tempered softmax consistently outperforms the BE-SNNs using normal softmax (labelled as BE-SNN-w/o in the figure), implying that the proposed adaptively tempered softmax can indeed improve the OOD detection performance of BE-SNNs.

\begin{figure}[H]
			\begin{center}
				\begin{tabular}{ccc}
					\includegraphics[width=0.3\linewidth]{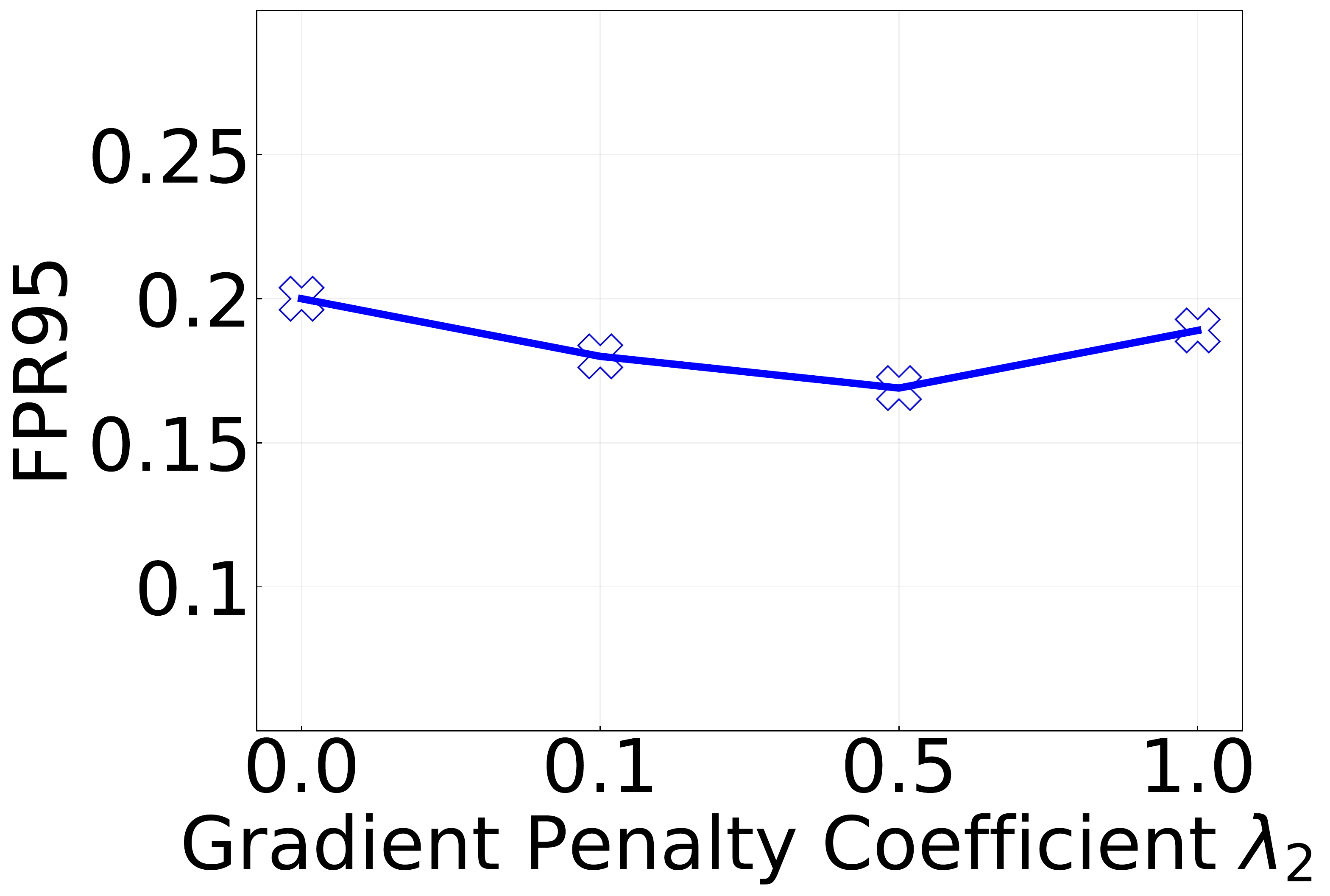}&\includegraphics[width=0.3\linewidth]{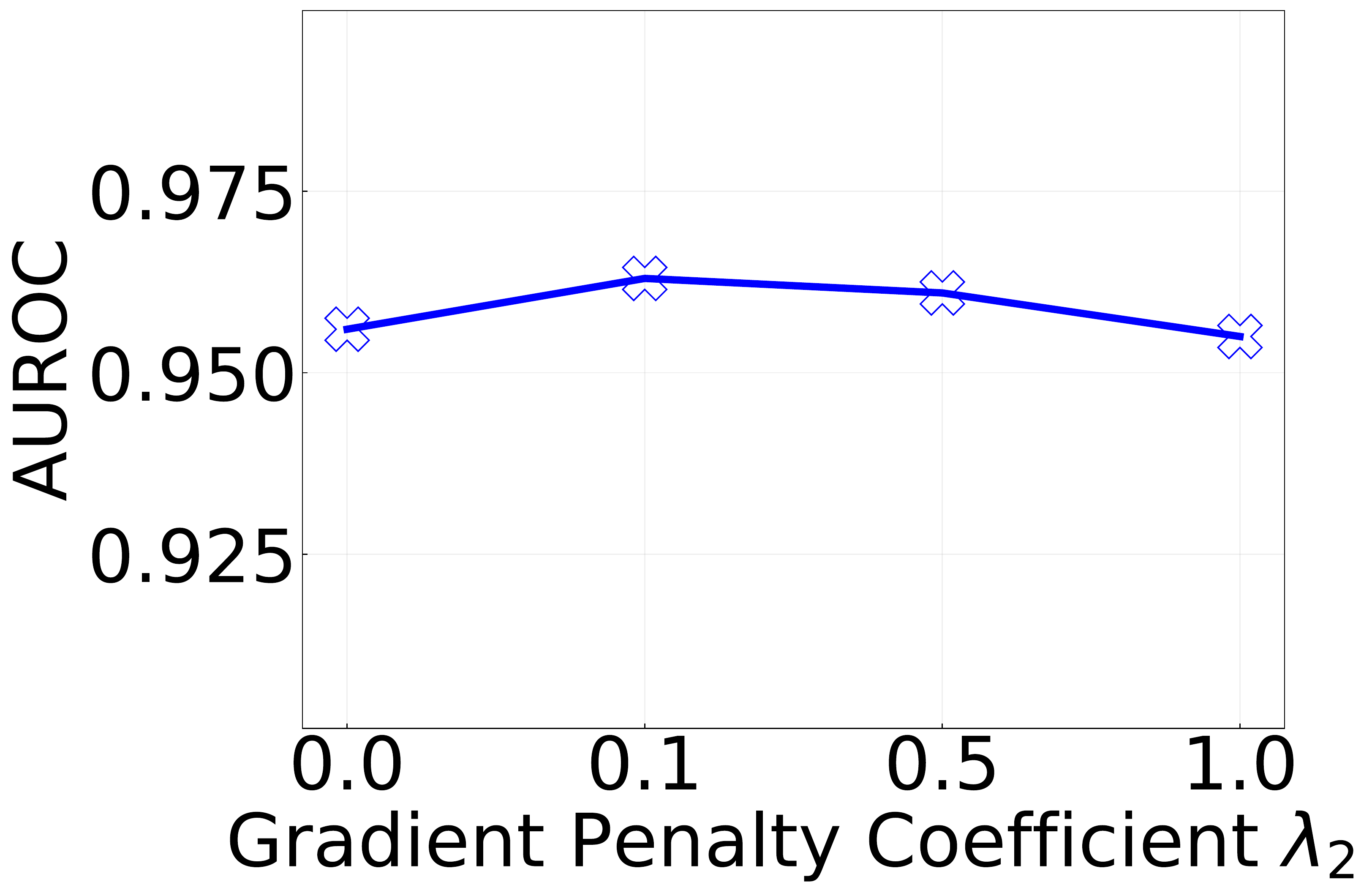} & \includegraphics[width=0.3\linewidth]{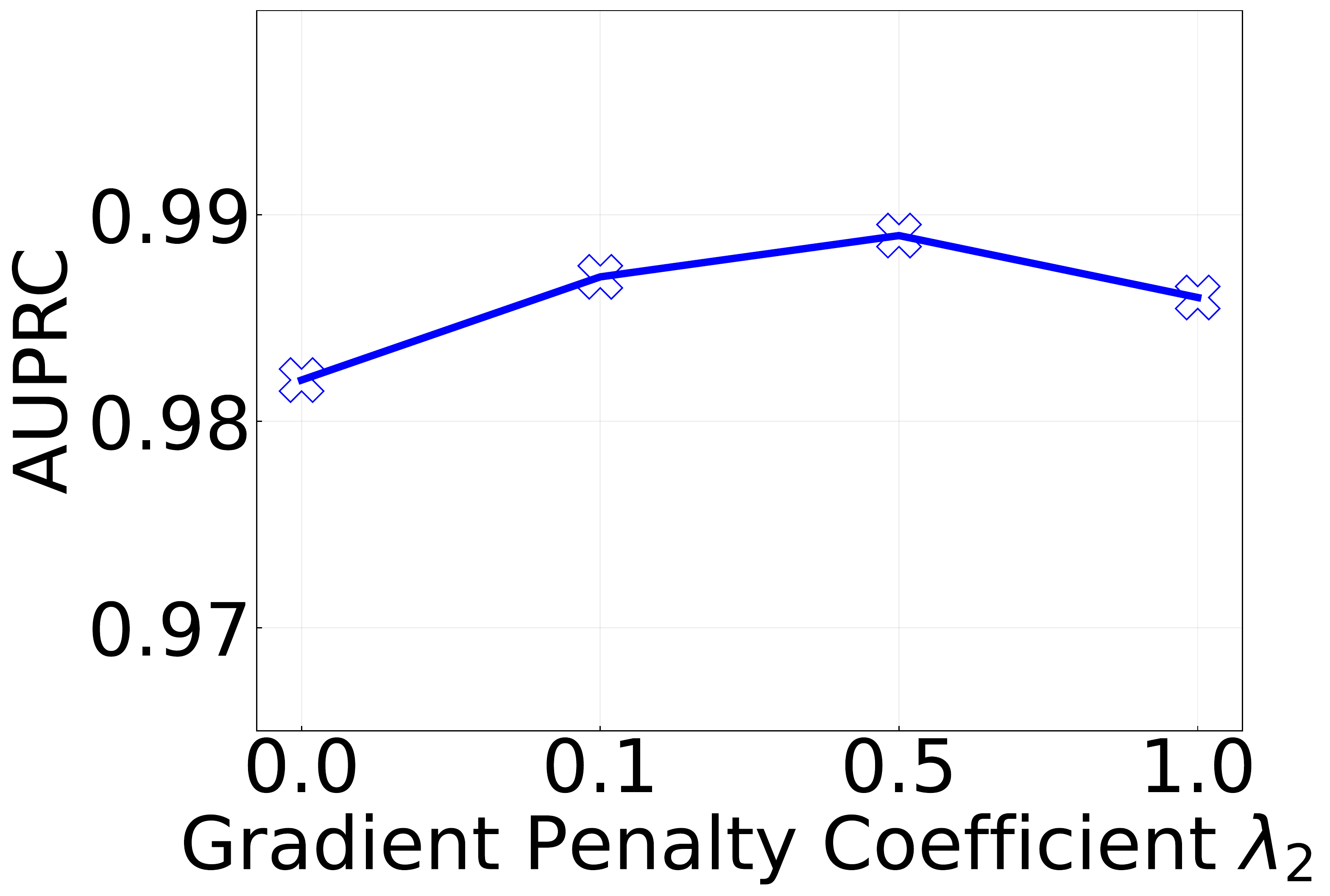}\\
					(a) FPR95 &(b) AUROC & (c) AUPRC\\

				\end{tabular}
				\caption{Comparisons of the OOD detection performance between BE-SNNs with different values of gradient penalty coefficient $\lambda_2$ in Equation~(\ref{eq:loss_snn}). Horizontal axis is the value of the coefficient $\lambda_2$, and the vertical axis is OOD detection metric. Lower FPR95, higher AUROC and AUPRC indicate better performance.}
				\label{figure:gp}
			\end{center}
		\end{figure}

		\textbf{Gradient penalty:} We present in Figure~\ref{figure:gp} the performance of BE-SNNs ($N_e=4$) with different values of gradient penalty coefficient $\lambda_2$ in Equation~(\ref{eq:loss_snn}). In particular, the regularization coefficient $\lambda_2$ is selected from the set \{0.0, 0.1, 0.5, 1.0\}. We found that the best performance of BE-SNNs occurs when $\lambda_2=0.5$, it can be observed from Figure~\ref{figure:gp} that the performance of BE-SNNs is not sensitive to the value of $\lambda_2$ in the set. In addition, we found in general the gradient penalty can enhance the ability of BE-SNNs to detect OOD samples, since all positive values of $\lambda_2$ result in better performance of BE-SNNs than $\lambda_2=0.0$.

		\label{appendix:ablation}
				
	\end{appendices}
	
\end{document}